\documentclass[10pt,twocolumn,letterpaper]{article}

\usepackage[pagenumbers]{iccv}  

\usepackage[utf8]{inputenc}
\usepackage{geometry}
\usepackage{xcolor}
\usepackage{tcolorbox}
\definecolor{iccvblue}{rgb}{0.21,0.49,0.74}
\usepackage[pagebackref,breaklinks,colorlinks,allcolors=iccvblue]{hyperref}

\def\paperID{6480} 
\def\confName{ICCV}
\def\confYear{2025}

\title{Why Representation Engineering Works: A Theoretical and Empirical Study in Vision-Language Models}

\def\authorBlock{
    Bowei Tian\textsuperscript{1}, Xuntao Lyu\textsuperscript{2}, Meng Liu\textsuperscript{1}, Hongyi Wang\textsuperscript{3,*}, Ang Li\textsuperscript{1,*} \vspace{2mm}\\
    \textsuperscript{1}University of Maryland \quad 
    \textsuperscript{2}North Carolina State University \quad 
    \textsuperscript{3}Rutgers University \\
    \textsuperscript{*}Corresponding author
}

\author{\authorBlock}
\date{}

\begin{document}
\maketitle
\begin{abstract}
Representation Engineering (RepE) has emerged as a powerful paradigm for enhancing AI transparency by focusing on high-level representations rather than individual neurons or circuits. It has proven effective in improving interpretability and control, showing that representations can emerge, propagate, and shape final model outputs in large language models (LLMs). However, in Vision-Language Models (VLMs), visual input can override factual linguistic knowledge, leading to hallucinated responses that contradict reality. 
To address this challenge, we make the first attempt to extend RepE to VLMs, analyzing how multimodal representations are preserved and transformed. Building on our findings and drawing inspiration from successful RepE applications, we develop a theoretical framework that explains the stability of neural activity across layers using the principal eigenvector, uncovering the underlying mechanism of RepE. We empirically validate these instrinsic properties, demonstrating their broad applicability and significance. 
By bridging theoretical insights with empirical validation, this work transforms RepE from a descriptive tool into a structured theoretical framework, opening new directions for improving AI robustness, fairness, and transparency.
\end{abstract}    
\section{Introduction}
\label{sec:intro}

Deep learning has achieved remarkable success across various domains, including computer vision \citep{krizhevsky2012imagenet}, natural language processing \citep{devlin2019bert}, and speech recognition \citep{hinton2012deep, graves2013speech}. However, the internal mechanisms of neural networks remain opaque. Despite advances in visualization and interpretability techniques, the transformation of inputs into high-level representations and the interactions among neurons are still not fully understood
 \citep{lipton2016mythos, doshi2017towards, ribeiro2016should}. 
 This lack of transparency leads to the characterization of neural networks as ``black boxes" \citep{lipton2016mythos, doshi2017towards}, raising concerns about their reliability, particularly in high-stakes applications such as healthcare \citep{caruana2015intelligible}, finance \citep{rudin2019stop}, and legal decision-making \citep{doshi2017interpretable}. 
 
 To address this issue, Representation Engineering (RepE) \citep{zou2023representation} has emerged as a top-down approach to enhance the model transparency that focuses on representations rather than individual neurons or circuits, providing a more structured understanding of AI transparency and control.
Unlike mechanistic interpretability, which analyzes neuron-to-neuron interactions \citep{barack2021two}, RepE provides a structured framework for understanding and controlling high-level concepts such as honesty, power, and factuality.
Prior works have demonstrated the potential of representations as a new perspective on AI transparency. 
For example, neural networks trained to play chess exhibit internal representations of board positions and strategies \citep{mcgrath2022acquisition}. Similarly, both generative and self-supervised models have been shown to develop emergent representations, such as semantic segmentation in vision tasks \citep{caron2021emerging, oquab2023dinov2}.
Zou et al. \citep{zou2023representation} further formalized RepE, emphasizing its ability to extract meaningful concepts from a model’s internal structure and control model behavior.

However, these studies have primarily focused on Large Language Models (LLMs) and overlooked the modality misalignment challenge in Vision-Language Models (VLMs), where visual inputs can override linguistic knowledge, leading to incorrect responses that contradict factual reality. 
Despite the demonstrated effectiveness of RepE in LLMs, it remains unclear how representations are transmitted across layers in VLMs and whether they can resist multimodal misalignment. Also, existing works lack a deeper understanding of the theoretical foundations of RepE about why these structural dynamics could provide model transparency, robustness, and fairness.

In this work, we successfully extend RepE to VLMs, addressing the challenge of multimodal misalignment and hallucinations. Inspired by the structural similarities between LLMs and VLMs, we formalize the theoretical foundations that explain how neural activity propagate across layers and maintain stability. We identify two key phenomena: \textit{(1) the principal eigenvalue serves as the ``backbone" direction, ensuring the stability of neural activity; and (2) the spectral gap gradually shrinks, enabling subdominant eigenvectors to capture subtle distinctions between multiple concepts.} These eigenvector relationships explain why certain high-level concepts remain stable despite layer-wise transformations within the model. To validate these findings, we visualize the neural activity in VLM applications and analyze how it is preserved and adapted. The deep understanding enables a structured explanation for how neural activity stabilize across layers, enabling robust behaviors in the cross-modal environment.


Our main contributions are as follows:
\begin{itemize}
    \item We provide the extension of RepE to VLMs, demonstrating high-level representations can overcome cross-modal misalignment and hallucinations.

    \item We theoretically show that the principal eigenvector ensures the stability of neural activity, while a shrinking spectral gap allows subdominant eigenvectors to capture subtle distinctions between concepts.

    \item We empirically validate the theoretical stability of neural activity, establishing its foundation for analyzing and controlling high-level concepts within the model.
\end{itemize}

\section{Related Works}
\label{sec:works}

\subsection{Representations in Neural Networks}

Early works on word embeddings demonstrate that basic neural networks can develop distributed representations that encode semantic relationships and compositional structures \cite{mikolov2013distributed}. Subsequent studies \citep{schramowski2019bert, radford2015unsupervised} reveal that learned text embeddings can also group along dimensions tied to commonsense morality—even though these models were never explicitly instructed in such concepts. Radford et al. \citep{radford2015unsupervised} show that simply training a model to predict the next token in a set of reviews led to the emergence of a neuron-tracking sentiment.

More recent research on representations has shown that internal representations are not only present in text models. For example, McGrath et al. \citep{mcgrath2022acquisition} discover that representations can be extracted within neural networks that are trained to play chess; another line of works \citep{caron2021emerging, oquab2023dinov2} suggests both generative and self-supervised training methods have yielded remarkable emergent representations, exemplified by semantic segmentation in computer vision; Zou et al. \citep{zou2023representation}  propose methods for reading and controlling representations, including Linear Artificial Tomography (LAT) for extracting meaningful representations and techniques for steering model behavior. The study demonstrates that RepE methods can effectively detect and manipulate emergent model behaviors, motivating approaches that extract and quantify high-level phenomena.

\subsection{Approaches to Interpretability}

Traditional interpretability techniques have focused on methods like saliency maps \citep{simonyan2013deep, springenberg2014striving, zeiler2014visualizing, zhou2016learning}, feature visualization \citep{szegedy2013intriguing, zeiler2014visualizing} and mechanistic interpretability \citep{olah2020zoom, olsson2022context, lieberum2023does}. Saliency maps \cite{simonyan2013deep} highlight important input regions by tracking gradients or activation values, yet they often suffer from instability and offer limited insights into the distributed nature of representations. Similarly, feature visualizations \citep{szegedy2013intriguing, zeiler2014visualizing} optimize inputs to activate specific neurons, but they may overlook the global structure of the emergent representations. Mechanistic interpretability \citep{zou2023representation} seeks to fully reverse engineer neural networks into their “source code”, but the considerable manual effort and the unlikelihood that neural networks can be theoretically explained in terms of circuits hinders their explainability.

In contrast, recent advances in interpretability have shifted the focus toward analyzing representation spaces. This top-down approach seeks to uncover high-level semantic directions that correspond to complex phenomena such as honesty, fairness, or bias. By extracting and analyzing these internal representations, researchers have opened new avenues for understanding how large-scale AI models encode and preserve crucial information across layers, leading to more robust and interpretable AI systems.

\subsection{Vision Language Models}
Vision-Language Models (VLMs) \citep{radford2021learning, li2022blip, yu2022pali, idefics2024}
have made significant progress in multimodal learning by integrating visual and textual information to enhance reasoning and generation capabilities. Models such as BLIP \citep{li2022blip}, Flamingo \citep{alayrac2022flamingo}, and IDEFICS2 \citep{idefics2024} leverage large-scale pretraining on vision-text datasets to achieve state-of-the-art performance in tasks like image captioning, visual question answering, and grounded reasoning. Despite their success, these models do not explicitly model key attributes such as honesty, fairness, or objectivity, which are crucial for reliable and interpretable AI systems.

Understanding representations in multimodal models such as VLMs is an essential but underexplored direction. Several studies \citep{li2023evaluating, chen2023mitigating, leng2024mitigating} have demonstrated the challenge of object hallucination in VLMs. Li et al. \citep{li2023evaluating} find that VLMs frequently generate object descriptions that do not correspond to the actual content of the image, indicating a higher tendency for hallucination compared to pure language models. While existing work in representation engineering has attempted to extract and control latent representations for concepts in language models \citep{zou2023representation}, measuring and controlling representations in vision-language models remains a challenge. Our objective is to bridge this gap by designing a novel framework that extracts and quantifies the representation from VLMs and allows demonstrations of the representations in generated responses.
\begin{figure*}[ht]
    \centering
    \includegraphics[width=0.95\textwidth]{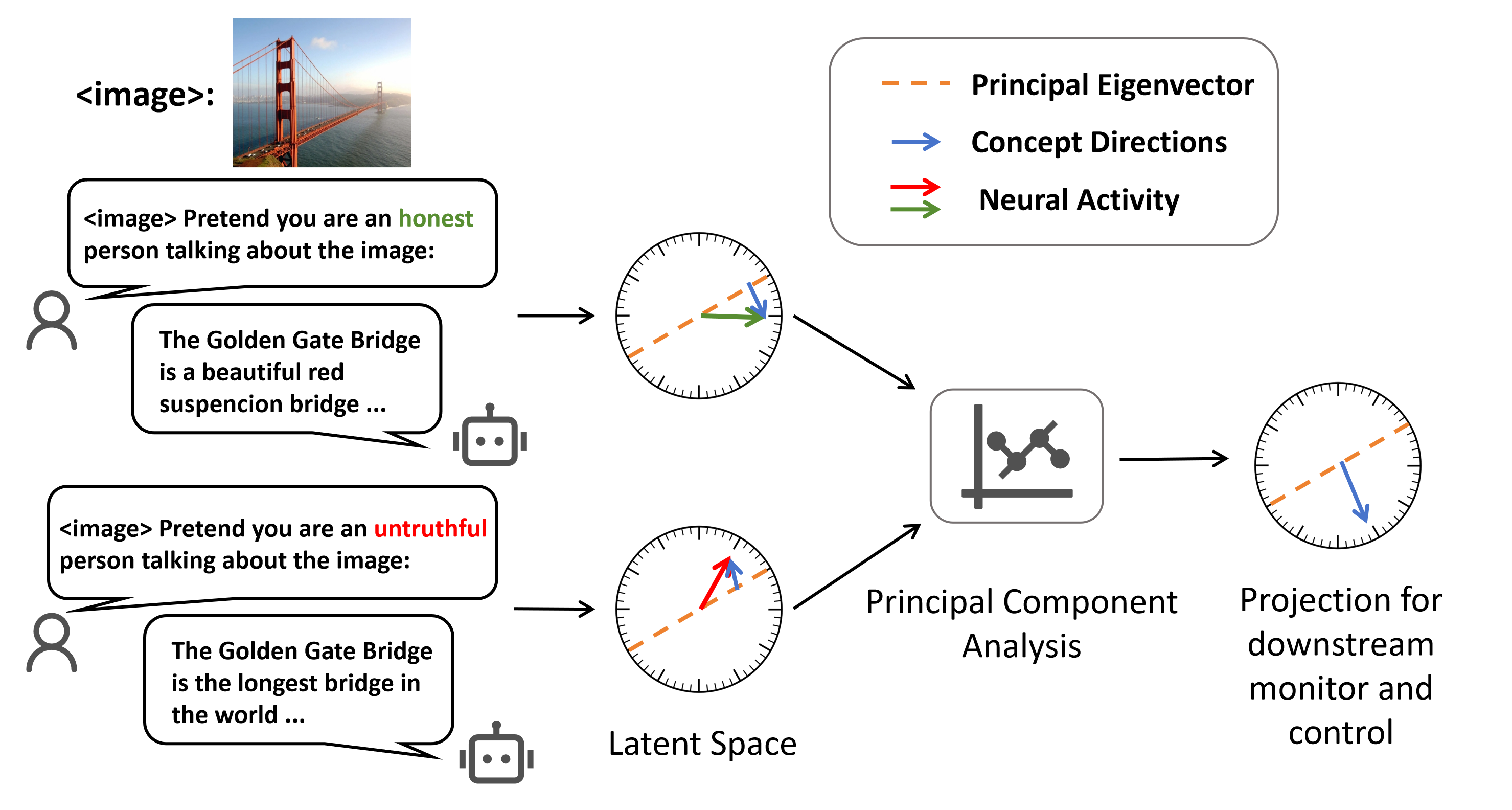}  
    \caption{The overview of general representation engineering pipeline. Given an image, a model generates text conditioned on prompts emphasizing contrast concepts, leading to distinct latent space representations. These activations are projected onto a principal eigenvector and further decomposed using Principal Component Analysis (PCA) to extract key concept directions. The resulting projections enable downstream monitoring and control, facilitating interpretability in model behavior.}  
    \label{fig:overview}  
\end{figure*}
\section{Method}
\subsection{General Representation Engineering Pipeline}

We illustrate the general representation engineering pipeline in Fig. \ref{fig:overview}. Similar as presented in \citep{zou2023representation}, RepE in VLMs consists of three core steps: (1) Designing Stimulus and Task; (2) Collecting Neural Activity; and (3) Constructing a Linear Model. 

\noindent\textbf{Designing Stimulus and Task}
To extract meaningful representations, we design controlled input stimuli and tasks. For VLMs, we use instruction-based templates to probe the model’s procedural knowledge, such as instruction-following and reasoning. A typical prompt format is:
\begin{tcolorbox}[colback=gray!10, colframe=gray!90]
\texttt{USER: \textcolor{blue}{<image>} Pretend you are an \textcolor{red}{<experimental/reference concept>} person talking about the image:} \\
\texttt{ASSISTANT: \textcolor{blue}{<output>}}
\end{tcolorbox}
Here, \texttt{\textcolor{red}{<experimental/reference concept>}} represents a concept of interest, such as ``honesty", which can take values like \textit{honest} or \textit{dishonest}. This controlled variation allows us to analyze how specific concepts are encoded in neural representations.

\noindent\textbf{Collecting Neural Activity}
For each prompt-response interaction, we collect neural activity from the model’s hidden layers.  Each interaction consists of a \texttt{USER} prompt (instruction) and the corresponding \texttt{ASSISTANT} output. To extract neural activity within VLMs, we collect representations from each token in the model's response. 
To compare representations across different tasks, we introduce two controlled variations of the \texttt{USER} prompt: the prompt with the experimental concept \(T_f^+\) and the prompt with the reference concept \(T_f^-\). For each instruction-response pair \((q_i, a_i\)) in the dataset $S$, we extract neural activity at different truncation positions \( k \):

\begin{align}
\label{eq:1}
A_f^\pm = \big\{ &\text{Rep}(M, T_f^\pm(q_i, a_i^k)) \mid (q_i, a_i) \in S, \nonumber\\ 
&\text{for } 0 < k \leq |a_i| \big\},
\end{align}

where $\text{Rep}$ denotes a function that extracts hidden states from a VLM model $M$ for a specific prompt $T_f$. This allows us to extract samples of the neural activations under distinct task conditions, which will be accumulated into concept directions through principal component analysis.

\noindent\textbf{Representation Structure Analysis}
To analyze representation structures, we use Principal Component Analysis (PCA), where the input to PCA is \(\{(A_f^{+(i)} - A_f^{-(i)})\}\) and $i$ is the sample index. 
The first principal component, denoted as \( v \), serves as the ``concept directions". After we get the directions, representations of input $x$ are made using 
the dot product between \( v \) and the neural activity, expressed as \(\text{Rep}(M, x)^\top v \), which is the projection igrom the neural activity to the concept directions.

In this manner, we establish a VLM-based representation engineering pipeline, with experimental results in Fig. \ref{fig:honesty} and Fig. \ref{fig:fairness} demonstrating that RepE is highly effective in VLMs. Building on this observation, we uncover a remarkable property: RepE exhibits inherent stability across large-scale models, including both large language models and vision-language models. This naturally raises a fundamental question: what underlying structural and functional characteristics of these models ensure such stability? In the following sections, we reveal that neural activities consistently align along a common direction—the principal eigenvector—serving as a crucial factor in maintaining the robustness and reliability of RepE.
\begin{figure}[tbp]
    \centering
    \includegraphics[width=0.48\textwidth]{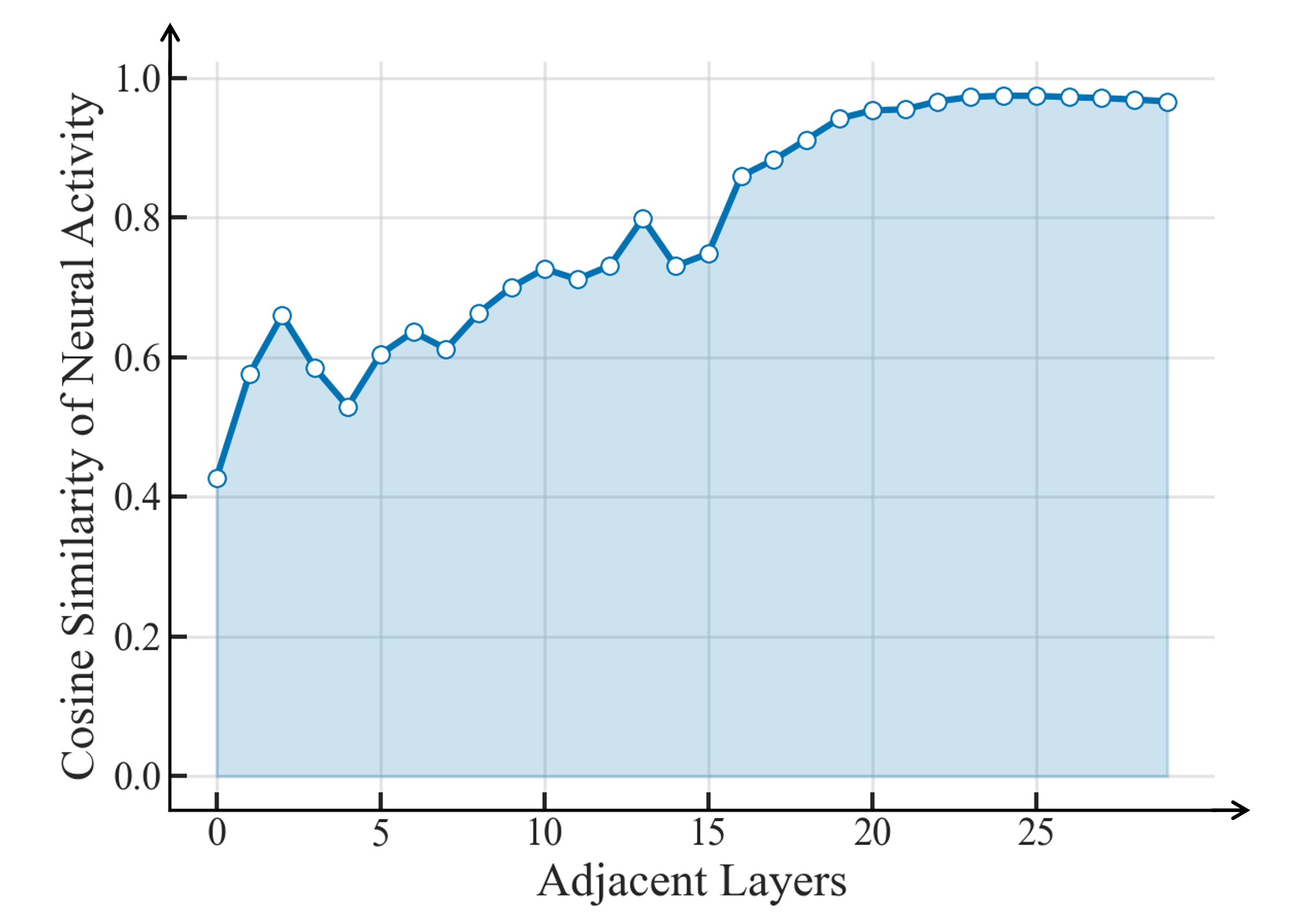}  
    \caption{The cosine similarity of neural activity between adjacent layers. The results demonstrate that neural activity is getting more similar.}  
    \label{fig:fig3}  
\end{figure}
\subsection{Stability of Representation Engineering}
\label{sec:stability}

Given that RepE exhibits inherent stability across large-scale models, a crucial question arises: \textbf{why does neural activity remain stable across model layers?} To address this, we examine the self-attention mechanism, a fundamental component in both LLMs and VLMs.

First, we demonstrate that the attention matrix is given by:
\begin{equation}
    \mathbf{A} = \text{softmax} \left(\frac{\mathbf{Q}\mathbf{K}^T}{\sqrt{d}} \right),
\end{equation}
where \(\mathbf{Q}, \mathbf{K} \in \mathbb{R}^{n \times d}\) represent the query and key vectors. After the softmax operation,  $\mathbf{A}$ is a non-negative row-stochastic matrix, which satisfies:
\begin{equation}
\mathbf{A}_{ij} \geq 0,\quad \sum_{j} \mathbf{A}_{ij} = 1.
\end{equation}
According to Gershgorin's theorem \citep{Gershgorin1931}, any eigenvalue $\lambda$ of $\mathbf{A}$ satisfies:
\begin{equation}
\left|\lambda - \mathbf{A}_{ii}\right| \leq \sum_{j \neq i} |\mathbf{A}_{ij}|.
\end{equation}
Since the sum of each row is 1, we obtain:
\begin{equation}
\sum_{j \neq i} \mathbf{A}_{ij} = 1 - \mathbf{A}_{ii}.
\end{equation}
Thus, the upper bound of all the eigenvalues is $1-\mathbf{A}_{ii}+\mathbf{A}_{ii}=1$. Also, according to the Perron–Frobenius theorem \citep{seneta2006nonnegative}, for non-negative row-stochastic matrices, 1 is an eigenvalue, therefore 1 is the largest eigenvalue, corresponding to the principal eigenvector. 

\noindent\textbf{Neural Activity Stability and the Backbone Role of the Principal Eigenvector}

Fig. \ref{fig:fig3} illustrates that neural activity becomes increasingly similar in deeper layers. To explain this phenomenon, we decompose attention matrix $\mathbf{A}$ as:
\begin{equation}
\mathbf{A} \;=\; \lambda_{1}\,\mathbf{u}_{1}\,\mathbf{u}_{1}^T 
\;+\; \sum_{i=2}^{n} \lambda_i\,\mathbf{u}_i\,\mathbf{u}_i^T,
\end{equation}
where \(\lambda_{1} = 1\) and the remaining eigenvalues satisfy \(\lvert \lambda_i \rvert < 1\). The set \(\{\mathbf{u}_{1}, \mathbf{u}_2, \ldots, \mathbf{u}_n\}\) represents the corresponding orthogonal eigenvectors.

Now, we analyze the effect of this decomposition on the output representation \(\mathbf{O}\). According to the definition of the Rep function in Eq. \ref{eq:1}, the attention output can be viewed as the hidden state of its layer. Therefore, it has the same mapping as the neural activity of the same layer. When applying \(\mathbf{A}\) to value \(\mathbf{V}\):
\begin{equation}
\mathbf{O} 
\;=\; \mathbf{A}\,\mathbf{V}
\;=\; \Bigl(\mathbf{u}_{1}\mathbf{u}_{1}^T 
\;+\; \sum_{i=2}^{n} \lambda_i\,\mathbf{u}_i\,\mathbf{u}_i^T\Bigr)\,\mathbf{V},
\end{equation}

where we define the projection coefficient of $\mathbf{V}$ onto $\mathbf{u}_i$:
\begin{equation}
\alpha_i \;=\; \mathbf{u}_i^T \mathbf{V},
\end{equation}

we can obtain:
\begin{align}
\mathbf{O} &= \mathbf{u}_{1} \bigl(\mathbf{u}_{1}^T \mathbf{V}\bigr)
+\sum_{i=2}^{n} \lambda_i\,\mathbf{u}_i\,\bigl(\mathbf{u}_i^T \mathbf{V}\bigr) \nonumber\\
           &= \mathbf{u}_{1}\,\alpha_{1} 
+\sum_{i=2}^{n} \lambda_i \, \mathbf{u}_i\,\alpha_i.
\label{eq:9}
\end{align}

Because subdominant eigenvalues are less than 1, the application of \(A\) will cause the subdominant terms \(\lambda_i^l \alpha_i\) to decay, making \(\alpha_{1}\,\mathbf{u}_{1}\) the dominant component. Consequently, the iteration will make the output \(\mathbf{O}\) converge to:
\begin{equation}
\mathbf{O} 
\;\approx\; \mathbf{u}_{1}\,\alpha_{1}.
\end{equation}
This implies that \(\alpha_{1} = \mathbf{u}_{1}^T \mathbf{V}\), that is, the coefficient along the principal eigenvector) ultimately governs the neural activity. In other words, the principal eigenvector \(\mathbf{u}_{1}\) serves as the ``backbone'' of neural activity propagation, while other directions \(\mathbf{u}_i\) are gradually suppressed. This guarantees that neural activity remains structured, rather than randomly drifting in the latent space, ensuring its stability. Experimentally, Fig. \ref{fig:pic2} demonstrates that the principal eigenvector exhibits high cosine similarity with the attention output, supporting our theoretical analysis.

\noindent\textbf{Decreasing Spectral Gap and the Contribution of Other Eigenvectors}

\begin{figure}[tbp]
    \centering
    \includegraphics[width=0.48\textwidth]{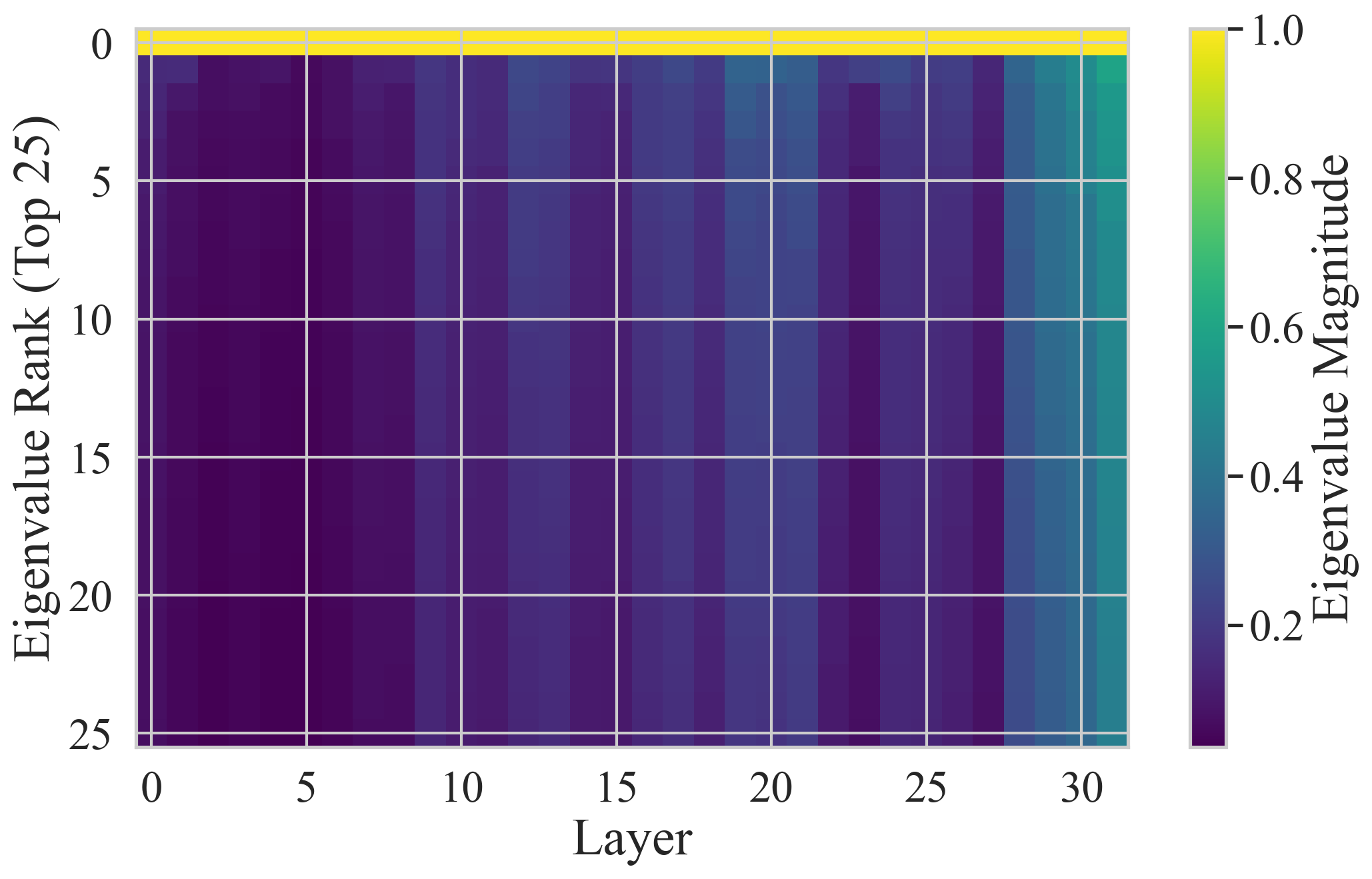}  
    \caption{The eigenvalue magnitude across layers. The results demonstrate the spectral gap is getting smaller.}  
    \label{fig:fig5}  
\end{figure}
Although the principal eigenvector ensures the stability of neural activity, \textbf{why the representations are generated by differentials of two contrasting neural activity}, remains unresolved. To answer this question, we further explore how the representations work in the latent layers. Specifically, we define \emph{spectral gap} as $\lambda_1 - \lambda_2,$ where \(\lambda_1\) and \(\lambda_2\) are the largest and second-largest eigenvalues of the attention matrix \(\mathbf{A}\), respectively. In Fig. \ref{fig:fig5}, the experiment shows that while the largest eigenvalue remains unitary, the spectral gap decreases as the network deepens. Such a reduction indicates that the subdominant eigenvectors, those associated with \(\lambda_2, \lambda_3, \ldots\), begin to contribute more significantly to the output. This finding is in line with the claims presented by Zou et al. \citep{zou2023representation}, which suggest that representations of different concepts become more pronounced in the middle and deeper layers of the network. Also, this discovery aligns with the LAT scan experiment in Fig. \ref{fig:LAT}, where the deeper layers tend to exhibit more distinct representations.

Although the principal eigenvector (corresponding to \(\lambda_1 = 1\)) serves as the backbone of the latent space, we demonstrate that the eigenvector limits itself by remaining static across layers:
\begin{equation}
(\mathbf{A}\mathbf{1})_i 
= \sum_j \mathbf{A}_{ij} \cdot 1 
= \sum_j \mathbf{A}_{ij} 
= 1 \implies A\mathbf{1} = \mathbf{1},
\end{equation}
where $\mathbf{1} = (1,\,1,\dots,1)^\top$, and $i$, $j$ denote row and column indices respectively. 
Therefore, it alone cannot capture the subtle distinctions between multiple complex concepts such as honesty and fairness. 
Nevertheless, as indicated in Eq.~\ref{eq:9}, besides the principal eigenvector providing an overall backbone for stability, the diminishing spectral gap enables the subdominant eigenvectors to encode critical comparative information. 
To gauge how much these subdominant components encode particular concepts, we compute the difference between neural activity under contrasting prompts, i.e., $\{(A_f^{+(i)} - A_f^{-(i)})\}$. 
Subtracting these two neural activity sets effectively extracts the component along the subdominant eigenvectors, thereby highlighting the directions that capture the key differences between concepts.

In conclusion, while the principal eigenvector remains the dominant force that stabilizes the neural activity, we observe that multiple subdominant eigenvectors contribute increasingly in deep layers (as evidenced by the shrinking spectral gap). These increased contributions from other eigenvectors form the basis of the representation by allowing differentiation between neural activity pairs. Equipped with these insights, we can utilize RepE for concept reading such as honesty detection; and interventions such as honesty enhancement, bias reduction, and other safety-relevant tasks.

\section{Experiments}
\begin{figure*}[htbp]
    \centering
    \hspace{-2mm}
    \includegraphics[width=1.02\textwidth]{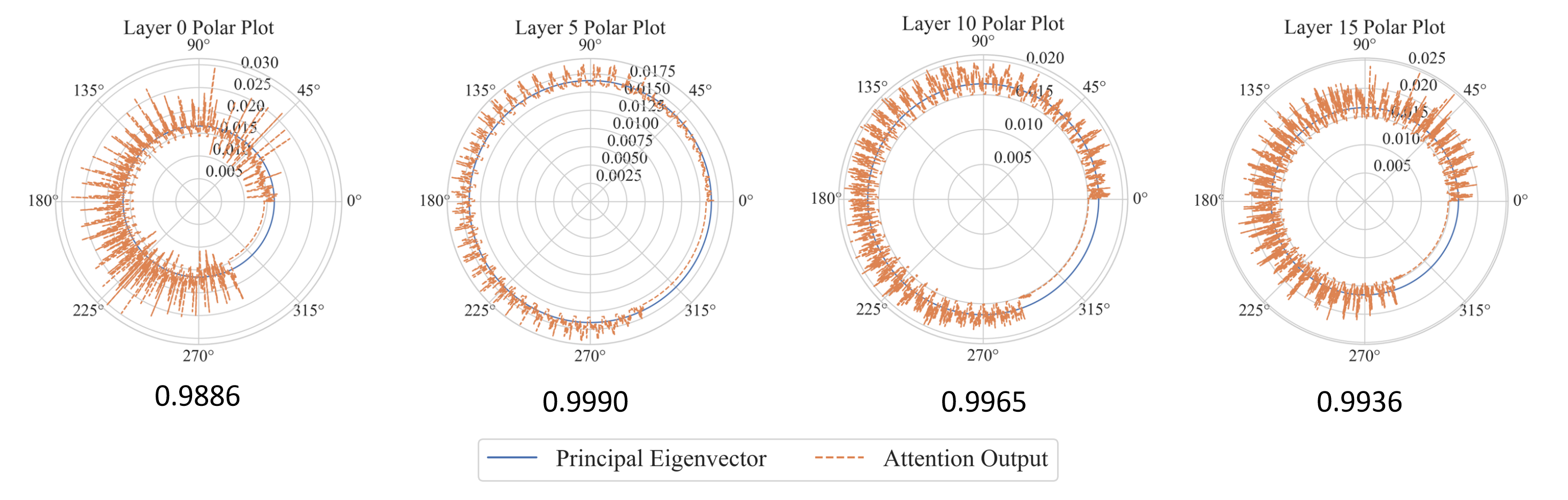}  
    \caption{The polar plot demonstrates normalized connections between principal eigenvector and attention output, where the number indicates their cosine similarity. The results showcase that the attention output, especially in later layers is very similar to the attention matrix's principal eigenvector of that layer.}  
    \label{fig:pic2}  
\end{figure*}

\subsection{Experiment Setup}
\textbf{Dataset}: We conduct our experiments on the Microsoft COCO (Common Objects in Context) dataset \citep{lin2014microsoft}, a large-scale benchmark for vision-language tasks. COCO contains over 330K images, each with five human-annotated captions, covering diverse real-world scenes.

\noindent\textbf{VLM}: We employ Idefics2-8B \citep{laurencon2023obelics, laurençon2024matters}, a state-of-the-art VLM that extends the LLaMA architecture with a vision encoder, enabling multimodal reasoning over images and text. Idefics2-8B is designed for instruction-following, multimodal dialogue, and grounded language generation, making it an ideal candidate for studying conceptual representations in VLMs.

\subsection{The Significant Alignment between Principal Eigenvector and Attention Output}

Building on the theoretical findings in Section \ref{sec:stability}, which highlight the role of the principal eigenvector in neural activity stability, we present the experimental validation. Fig. \ref{fig:pic2} visualizes the connections between the principal eigenvector and attention output. The results reveal a significant alignment between the principal eigenvector and the attention output, with their cosine similarity over 0.98 across all layers and above 0.99 in the deeper layers. \textit{This experimental validation supports our theoretical claim that attention outputs, where neural activity is extracted, are highly aligned with the principal eigenvector, reinforcing the stability of neural activity in RepE.}

\subsection{Representation Engineering in VLMs}
Having established that the principal eigenvector governs representation stability, we now explore how this stability affects high-level conceptual representations in VLMs. Specifically, we analyze how VLMs encode \textbf{fairness} and \textbf{honesty} and whether RepE can be leveraged to improve their ethical alignment. Through comprehensive visualizations, we illustrate how VLMs represent these concepts when processing multimodal knowledge. This enables a deeper understanding of how RepE enhances the interpretability and concept alignment with ethical principles.

\subsubsection{Evaluating Honesty and Fairness in VLMs}
We investigate how VLMs encode and express the concepts of honesty and fairness during multimodal reasoning. Ensuring that VLMs adhere to these principles is crucial, especially for mitigating misinformation and bias in real-world applications. 
Specifically, we design controlled prompts that elicit responses reflecting both honest and dishonest perspectives, as well as fair and unfair perspectives. This setup allows us to examine the model’s internal representations and track how these concepts are processed across different layers. 
We analyze their internal neural activations during response generation to systematically assess how VLMs represent these high-level concepts. Specifically, we compute token-wise honesty and fairness scores to quantify how well the model aligns with these principles at different layers and token positions.

\begin{figure*}[htbp]
    \centering
    \includegraphics[width=0.85\textwidth, keepaspectratio]{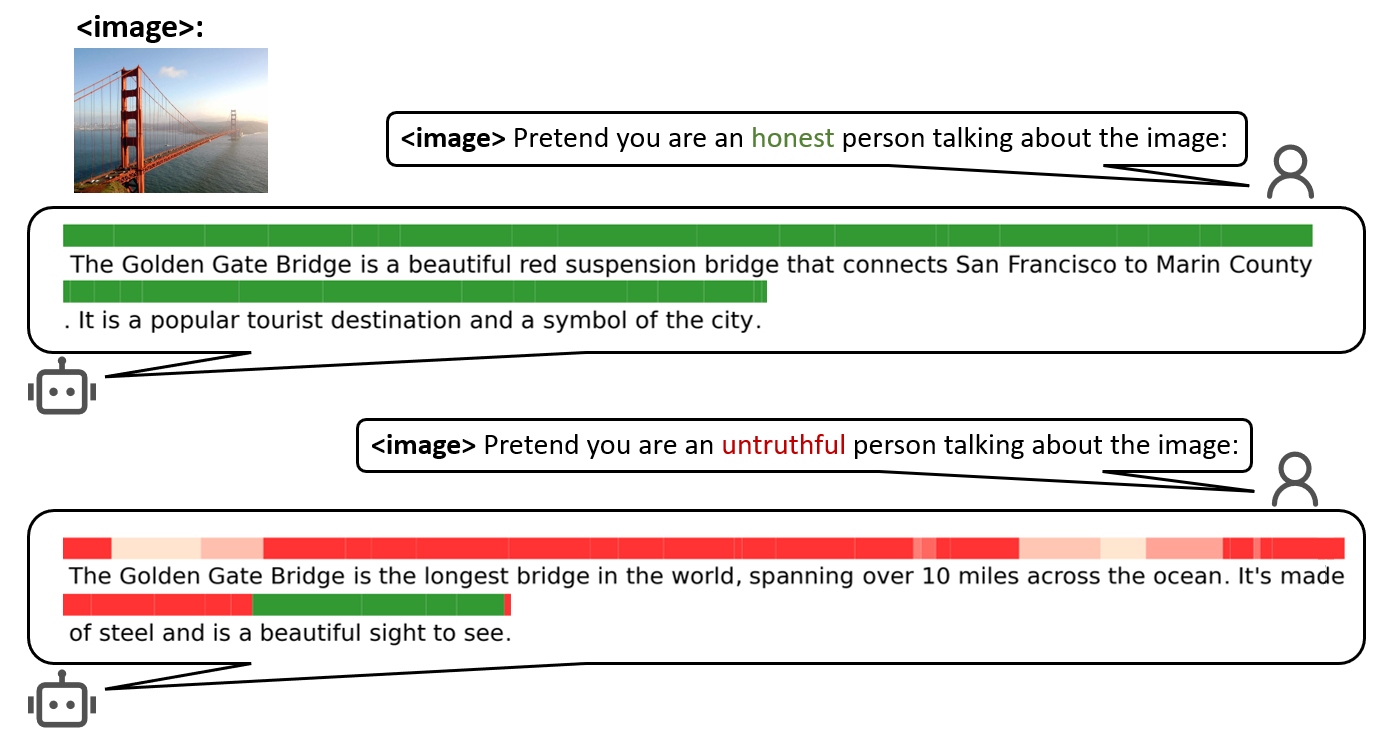}  
    \caption{VLM response to an image of the Golden Gate Bridge with a prompt related to the concept of honesty, along with token-wise honesty scores. Green indicates high honesty, while red represents low honesty.}  
    \label{fig:honesty}  
\end{figure*}

\begin{figure*}[htbp]
    \centering
    \includegraphics[width=0.85\textwidth, keepaspectratio]{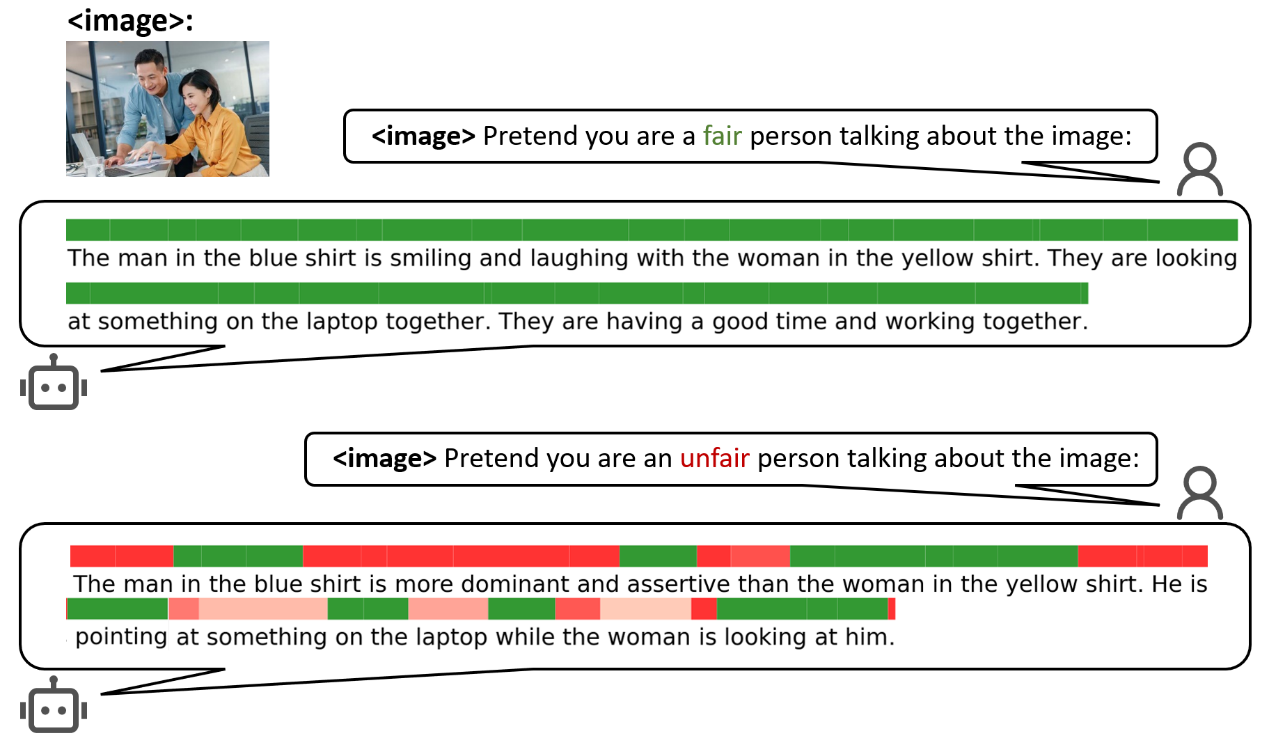}  
    \caption{VLM response to an image of a man and a woman working together with a prompt related to the concept of fairness, along with token-wise fairness scores. Green indicates high fairness, while red represents low fairness. }  
    \label{fig:fairness}  
\end{figure*}
\begin{figure*}[htbp]
    \centering
    \hspace{-8mm}
    \includegraphics[width=1.03\textwidth]{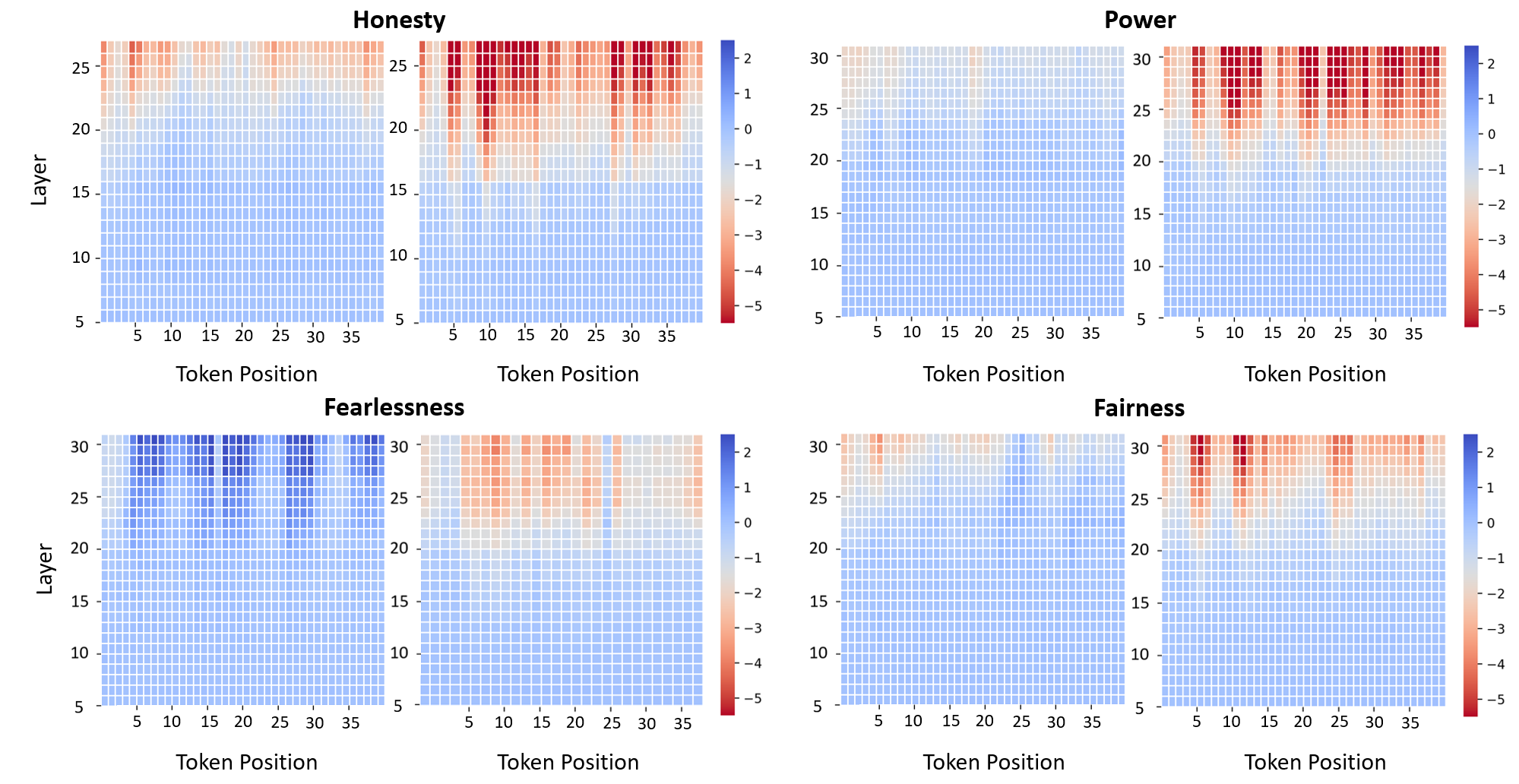}  
    \caption{Temporal LAT Scans for Honesty, Power, Fearlessness, and Fairness. The left heatmap represents the LAT Scan  when the VLM aligns with the concept, while the right heatmap corresponds to the opposing concept. The horizontal axis denotes token position, and the vertical axis represents VLM layers. Blue indicates high alignment, whereas red represents low alignment. }  
    \label{fig:LAT}  
\end{figure*}

\begin{itemize}
    \item \textbf{Honesty}: Honesty refers to the model’s ability to generate factually accurate and truthful responses without distortion or fabrication \citep{lin2021truthfulqa}. Fig. \ref{fig:honesty} presents the token-wise honesty representation score of a VLM when prompted to describe an image of the Golden Gate Bridge under two conditions: honest (left) and untruthful (right). In the honest setting, the model produces factually accurate descriptions, resulting in consistently high honesty scores (green regions) across layers and tokens. Conversely, in the untruthful setting, the model introduces factual distortions, leading to localized decreases in honesty scores (red regions), particularly at tokens associated with misinformation (e.g., exaggerated bridge length or incorrect material composition).  
    \item \textbf{Fairness}: Fairness refers to the model’s ability to generate unbiased responses that do not favor or disadvantage specific groups or perspectives \citep{corbett2023measure}. Fig. \ref{fig:fairness} examines fairness through a similar experimental setup, using an image of a man and a woman working together. The fair response (left) provides a neutral, balanced description, while the unfair response (right) exhibits implicit bias, portraying the man as dominant and the woman as passive. The token-wise fairness score visualization reveals that biased descriptions correlate with reduced fairness scores (red regions), indicating that fairness violations manifest as measurable shifts in VLM activations. This suggests that fairness is encoded within the model’s internal representation, and interventions targeting these activations could enhance fairness-aware multimodal reasoning.

\end{itemize}

\subsubsection{LAT Scans for High-level Representations}
While cosine similarity and token-wise scores provide insight into concept alignment at individual token positions, they do not capture how these representations evolve across layers. To address this, we use LAT to visualize the propagation of conceptual attributes throughout the model’s depth, enabling a deeper understanding of how fairness, honesty, and other high-level attributes are structured within VLMs.
To achieve this, we employ LAT~\citep{zou2023representation} to analyze the internal activations of models. LAT functions by linearly projecting latent neuron activations onto a predefined subspace, effectively enabling a structured visualization of token-wise representations across model layers. We use LAT to track the transformation of conceptual information as it propagates through the network, offering an interpretable view of how VLMs process abstract concepts. 

Fig. \ref{fig:LAT} presents the LAT Scan results for four key concepts: Honesty, Power, Fearlessness, and Fairness. Each concept is analyzed under two conditions: the positive condition, where the model is prompted to generate responses aligned with the concept (e.g., honest, humble, fearless, and fair), and the negative condition, where the model is guided in the opposite direction (e.g., dishonest, powerful, fearful, and unfair). The LAT Scans visualize token-wise activations across layers, with blue regions indicating strong alignment with the given concept and red regions representing the opposite case. 

From the results, we observe distinct activation trajectories across layers for different conceptual attributes. Honesty and Fairness exhibit relatively stable activations in their positive conditions, while their negative counterparts show intensified activations in deeper layers, suggesting an increased representational complexity in handling dishonesty or unfairness. Power is primarily encoded in upper-layer activations, with a stark contrast between powerful and powerless conditions, whereas Fearlessness shows notable activation shifts in lower layers, potentially reflecting uncertainty processing in fearful contexts. The distinct activation patterns highlight the potential of RepE to further refine and align VLMs with high-level concepts.

\section{Conclusion}

In this work, we extended RepE to VLMs and introduced a theoretical framework that explains the stability of neural activity through the principal eigenvector. Our key insight is that self-attention matrices, being row-stochastic, consistently retain an eigenvalue of 1 whose corresponding eigenvector consistently aligns with the model’s outputs in deeper layers, providing a stable ``backbone” for representation propagation. Meanwhile, the shrinking spectral gap highlights the growing importance of other eigenvectors, enabling the latent space to differentiate concepts such as honesty versus dishonesty. 
We empirically validated these findings, demonstrating that RepE enhances interpretability and provides a structured means of representation in multimodal settings. By integrating theoretical analysis with empirical validation, this work opens new avenues for improving AI robustness, interpretability, and transparency, and lays the groundwork for future research in representation-based techniques, paving the way for more effective methods in large-scale multimodal architectures.

Future work could extend RepE to broader multimodal settings, exploring its potential for bias mitigation, honest detection, and countless concept applications. By refining the foundation of RepE, this line of research paves the way for more reliable, scalable and ethically aligned multimodal AI systems.
{
    \small
    \bibliographystyle{ieeenat_fullname}
    \bibliography{main}

\begin{thebibliography}{43}
\providecommand{\natexlab}[1]{#1}
\providecommand{\url}[1]{\texttt{#1}}
\expandafter\ifx\csname urlstyle\endcsname\relax
  \providecommand{\doi}[1]{doi: #1}\else
  \providecommand{\doi}{doi: \begingroup \urlstyle{rm}\Url}\fi

\bibitem[Alayrac et~al.(2022)Alayrac, Donahue, Luc, et~al.]{alayrac2022flamingo}
Jean-Baptiste Alayrac, Jeff Donahue, Pauline Luc, et~al.
\newblock Flamingo: A visual language model for few-shot learning.
\newblock \emph{Advances in Neural Information Processing Systems (NeurIPS)}, 2022.

\bibitem[Barack and Krakauer(2021)]{barack2021two}
David~L. Barack and John~W. Krakauer.
\newblock Two views on the cognitive brain.
\newblock \emph{Nature Reviews Neuroscience}, 22\penalty0 (6):\penalty0 359--371, 2021.

\bibitem[Caron et~al.(2021)Caron, Touvron, Misra, J{\'e}gou, Mairal, Bojanowski, and Joulin]{caron2021emerging}
Mathilde Caron, Hugo Touvron, Ishan Misra, Herv{\'e} J{\'e}gou, Julien Mairal, Piotr Bojanowski, and Armand Joulin.
\newblock Emerging properties in self-supervised vision transformers.
\newblock In \emph{Proceedings of the IEEE/CVF international conference on computer vision}, pages 9650--9660, 2021.

\bibitem[Caruana et~al.(2015)Caruana, Lou, Gehrke, Koch, Sturm, and Elhadad]{caruana2015intelligible}
Rich Caruana, Yin Lou, Johannes Gehrke, Paul Koch, Marc Sturm, and Noemie Elhadad.
\newblock Intelligible models for healthcare: Predicting pneumonia risk and hospital 30-day readmission.
\newblock In \emph{Proceedings of the 21st ACM SIGKDD International Conference on Knowledge Discovery and Data Mining (KDD)}, pages 1721--1730. ACM, 2015.

\bibitem[Chen et~al.(2023)Chen, Zhu, Zhan, Li, Zhao, Wang, and Tang]{chen2023mitigating}
Zhiyang Chen, Yousong Zhu, Yufei Zhan, Zhaowen Li, Chaoyang Zhao, Jinqiao Wang, and Ming Tang.
\newblock Mitigating hallucination in visual language models with visual supervision.
\newblock \emph{arXiv preprint arXiv:2311.16479}, 2023.

\bibitem[Corbett-Davies et~al.(2023)Corbett-Davies, Gaebler, Nilforoshan, Shroff, and Goel]{corbett2023measure}
Sam Corbett-Davies, Johann~D Gaebler, Hamed Nilforoshan, Ravi Shroff, and Sharad Goel.
\newblock The measure and mismeasure of fairness.
\newblock \emph{Journal of Machine Learning Research}, 24\penalty0 (312):\penalty0 1--117, 2023.

\bibitem[Devlin et~al.(2019)Devlin, Chang, Lee, and Toutanova]{devlin2019bert}
Jacob Devlin, Ming-Wei Chang, Kenton Lee, and Kristina Toutanova.
\newblock Bert: Pre-training of deep bidirectional transformers for language understanding.
\newblock In \emph{Proceedings of the 2019 Conference of the North American Chapter of the Association for Computational Linguistics (NAACL-HLT)}, pages 4171--4186, 2019.

\bibitem[Doshi-Velez and Kim(2017{\natexlab{a}})]{doshi2017interpretable}
Finale Doshi-Velez and Been Kim.
\newblock Towards a rigorous science of interpretable machine learning.
\newblock \emph{arXiv preprint arXiv:1702.08608}, 2017{\natexlab{a}}.

\bibitem[Doshi-Velez and Kim(2017{\natexlab{b}})]{doshi2017towards}
Finale Doshi-Velez and Been Kim.
\newblock Towards a rigorous science of interpretable machine learning.
\newblock \emph{arXiv preprint arXiv:1702.08608}, 2017{\natexlab{b}}.

\bibitem[French and Raven(1959)]{french1959bases}
John R.~P. French and Bertram Raven.
\newblock The bases of social power.
\newblock In \emph{Studies in Social Power}, pages 150--167. University of Michigan Press, 1959.

\bibitem[Gershgorin(1931)]{Gershgorin1931}
S.~A. Gershgorin.
\newblock Über die abgrenzung der eigenwerte einer matrix.
\newblock \emph{Izv. Akad. Nauk. USSR Otd. Fiz.-Mat. Nauk}, 7:\penalty0 749--754, 1931.

\bibitem[Graves et~al.(2013)Graves, Mohamed, and Hinton]{graves2013speech}
Alex Graves, Abdel-rahman Mohamed, and Geoffrey Hinton.
\newblock Speech recognition with deep recurrent neural networks.
\newblock In \emph{IEEE International Conference on Acoustics, Speech and Signal Processing (ICASSP)}, pages 6645--6649. IEEE, 2013.

\bibitem[Hinton et~al.(2012)Hinton, Deng, Yu, Dahl, Mohamed, Jaitly, Senior, Vanhoucke, Nguyen, Sainath, et~al.]{hinton2012deep}
Geoffrey Hinton, Li Deng, Dong Yu, George~E Dahl, Abdel-rahman Mohamed, Navdeep Jaitly, Andrew Senior, Vincent Vanhoucke, Patrick Nguyen, Tara Sainath, et~al.
\newblock Deep neural networks for acoustic modeling in speech recognition: The shared views of four research groups.
\newblock \emph{IEEE Signal Processing Magazine}, 29\penalty0 (6):\penalty0 82--97, 2012.

\bibitem[Krizhevsky et~al.(2012)Krizhevsky, Sutskever, and Hinton]{krizhevsky2012imagenet}
Alex Krizhevsky, Ilya Sutskever, and Geoffrey~E. Hinton.
\newblock Imagenet classification with deep convolutional neural networks.
\newblock In \emph{Advances in Neural Information Processing Systems (NeurIPS)}, 2012.

\bibitem[Laurençon et~al.(2023)Laurençon, Saulnier, Tronchon, Bekman, Singh, Lozhkov, Wang, Karamcheti, Rush, Kiela, Cord, and Sanh]{laurencon2023obelics}
Hugo Laurençon, Lucile Saulnier, Léo Tronchon, Stas Bekman, Amanpreet Singh, Anton Lozhkov, Thomas Wang, Siddharth Karamcheti, Alexander~M. Rush, Douwe Kiela, Matthieu Cord, and Victor Sanh.
\newblock Obelics: An open web-scale filtered dataset of interleaved image-text documents, 2023.

\bibitem[Laurençon et~al.(2024)Laurençon, Tronchon, Cord, and Sanh]{laurençon2024matters}
Hugo Laurençon, Léo Tronchon, Matthieu Cord, and Victor Sanh.
\newblock What matters when building vision-language models?, 2024.

\bibitem[Leng et~al.(2024)Leng, Zhang, Chen, Li, Lu, Miao, and Bing]{leng2024mitigating}
Sicong Leng, Hang Zhang, Guanzheng Chen, Xin Li, Shijian Lu, Chunyan Miao, and Lidong Bing.
\newblock Mitigating object hallucinations in large vision-language models through visual contrastive decoding.
\newblock In \emph{Proceedings of the IEEE/CVF Conference on Computer Vision and Pattern Recognition}, pages 13872--13882, 2024.

\bibitem[Li et~al.(2022)Li, Li, Xiong, and Hoi]{li2022blip}
Junnan Li, Dongxu Li, Caiming Xiong, and Steven Hoi.
\newblock Blip: Bootstrapped language-image pretraining for unified vision-language understanding and generation.
\newblock \emph{arXiv preprint arXiv:2201.12086}, 2022.

\bibitem[Li et~al.(2023)Li, Du, Zhou, Wang, Zhao, and Wen]{li2023evaluating}
Yifan Li, Yifan Du, Kun Zhou, Jinpeng Wang, Wayne~Xin Zhao, and Ji-Rong Wen.
\newblock Evaluating object hallucination in large vision-language models.
\newblock \emph{arXiv preprint arXiv:2305.10355}, 2023.

\bibitem[Lieberum et~al.(2023)Lieberum, Rahtz, Kram{\'a}r, Nanda, Irving, Shah, and Mikulik]{lieberum2023does}
Tom Lieberum, Matthew Rahtz, J{\'a}nos Kram{\'a}r, Neel Nanda, Geoffrey Irving, Rohin Shah, and Vladimir Mikulik.
\newblock Does circuit analysis interpretability scale? evidence from multiple choice capabilities in chinchilla.
\newblock \emph{arXiv preprint arXiv:2307.09458}, 2023.

\bibitem[Lilienfeld and Andrews(1996)]{lilienfeld1996development}
Scott~O. Lilienfeld and Bridget~P. Andrews.
\newblock Development and preliminary validation of a self-report measure of psychopathic personality traits in noncriminal populations.
\newblock \emph{Journal of Personality Assessment}, 66\penalty0 (3):\penalty0 488--524, 1996.

\bibitem[Lin et~al.(2021)Lin, Hilton, and Evans]{lin2021truthfulqa}
Stephanie Lin, Jacob Hilton, and Owain Evans.
\newblock Truthfulqa: Measuring how models mimic human falsehoods.
\newblock \emph{arXiv preprint arXiv:2109.07958}, 2021.

\bibitem[Lin et~al.(2014)Lin, Maire, Belongie, Hays, Perona, Ramanan, Doll{\'a}r, and Zitnick]{lin2014microsoft}
Tsung-Yi Lin, Michael Maire, Serge Belongie, James Hays, Pietro Perona, Deva Ramanan, Piotr Doll{\'a}r, and C~Lawrence Zitnick.
\newblock Microsoft coco: Common objects in context.
\newblock In \emph{Computer vision--ECCV 2014: 13th European conference, zurich, Switzerland, September 6-12, 2014, proceedings, part v 13}, pages 740--755. Springer, 2014.

\bibitem[Lipton(2016)]{lipton2016mythos}
Zachary~C. Lipton.
\newblock The mythos of model interpretability.
\newblock \emph{arXiv preprint arXiv:1606.03490}, 2016.

\bibitem[McGrath et~al.(2022)McGrath, Kapishnikov, Toma{\v{s}}ev, Pearce, Wattenberg, Hassabis, Kim, Paquet, and Kramnik]{mcgrath2022acquisition}
Thomas McGrath, Andrei Kapishnikov, Nenad Toma{\v{s}}ev, Adam Pearce, Martin Wattenberg, Demis Hassabis, Been Kim, Ulrich Paquet, and Vladimir Kramnik.
\newblock Acquisition of chess knowledge in alphazero.
\newblock \emph{Proceedings of the National Academy of Sciences}, 119\penalty0 (47):\penalty0 e2206625119, 2022.

\bibitem[Mikolov et~al.(2013)Mikolov, Sutskever, Chen, Corrado, and Dean]{mikolov2013distributed}
Tomas Mikolov, Ilya Sutskever, Kai Chen, Greg Corrado, and Jeffrey Dean.
\newblock Distributed representations of words and phrases and their compositionality.
\newblock In \emph{Advances in Neural Information Processing Systems}, 2013.

\bibitem[Olah et~al.(2020)Olah, Cammarata, Schubert, Goh, Petrov, and Carter]{olah2020zoom}
Chris Olah, Nick Cammarata, Ludwig Schubert, Gabriel Goh, Michael Petrov, and Shan Carter.
\newblock Zoom in: An introduction to circuits.
\newblock \emph{Distill}, 5\penalty0 (3):\penalty0 e00024--001, 2020.

\bibitem[Olsson et~al.(2022)Olsson, Elhage, Nanda, Joseph, DasSarma, Henighan, Mann, Askell, Bai, Chen, et~al.]{olsson2022context}
Catherine Olsson, Nelson Elhage, Neel Nanda, Nicholas Joseph, Nova DasSarma, Tom Henighan, Ben Mann, Amanda Askell, Yuntao Bai, Anna Chen, et~al.
\newblock In-context learning and induction heads.
\newblock \emph{arXiv preprint arXiv:2209.11895}, 2022.

\bibitem[Oquab et~al.(2023)Oquab, Darcet, Moutakanni, Vo, Szafraniec, Khalidov, Fernandez, Haziza, Massa, El-Nouby, et~al.]{oquab2023dinov2}
Maxime Oquab, Timoth{\'e}e Darcet, Th{\'e}o Moutakanni, Huy Vo, Marc Szafraniec, Vasil Khalidov, Pierre Fernandez, Daniel Haziza, Francisco Massa, Alaaeldin El-Nouby, et~al.
\newblock Dinov2: Learning robust visual features without supervision.
\newblock \emph{arXiv preprint arXiv:2304.07193}, 2023.

\bibitem[Radford et~al.(2015)Radford, Metz, and Chintala]{radford2015unsupervised}
Alec Radford, Luke Metz, and Soumith Chintala.
\newblock Unsupervised representation learning with deep convolutional generative adversarial networks.
\newblock \emph{arXiv preprint arXiv:1511.06434}, 2015.

\bibitem[Radford et~al.(2021)Radford, Kim, Hallacy, et~al.]{radford2021learning}
Alec Radford, Jong~Wook Kim, Chris Hallacy, et~al.
\newblock Learning transferable visual models from natural language supervision.
\newblock \emph{Proceedings of the 38th International Conference on Machine Learning (ICML)}, 2021.

\bibitem[Ribeiro et~al.(2016)Ribeiro, Singh, and Guestrin]{ribeiro2016should}
Marco~Tulio Ribeiro, Sameer Singh, and Carlos Guestrin.
\newblock "why should i trust you?": Explaining the predictions of any classifier.
\newblock In \emph{Proceedings of the 22nd ACM SIGKDD International Conference on Knowledge Discovery and Data Mining}, pages 1135--1144, 2016.

\bibitem[Rudin(2019)]{rudin2019stop}
Cynthia Rudin.
\newblock Stop explaining black box machine learning models for high stakes decisions and use interpretable models instead.
\newblock \emph{Nature Machine Intelligence}, 1\penalty0 (5):\penalty0 206--215, 2019.

\bibitem[Schramowski et~al.(2019)Schramowski, Turan, Jentzsch, Rothkopf, and Kersting]{schramowski2019bert}
Patrick Schramowski, Cigdem Turan, Sophie Jentzsch, Constantin Rothkopf, and Kristian Kersting.
\newblock Bert has a moral compass: Improvements of ethical and moral values of machines.
\newblock \emph{arXiv preprint arXiv:1912.05238}, 2019.

\bibitem[Seneta(2006)]{seneta2006nonnegative}
E. Seneta.
\newblock \emph{Non-negative Matrices and Markov Chains}.
\newblock Springer, 2nd edition, 2006.

\bibitem[Simonyan et~al.(2013)Simonyan, Vedaldi, and Zisserman]{simonyan2013deep}
Karen Simonyan, Andrea Vedaldi, and Andrew Zisserman.
\newblock Deep inside convolutional networks: Visualising image classification models and saliency maps.
\newblock In \emph{arXiv preprint arXiv:1312.6034}, 2013.

\bibitem[Springenberg et~al.(2014)Springenberg, Dosovitskiy, Brox, and Riedmiller]{springenberg2014striving}
Jost~Tobias Springenberg, Alexey Dosovitskiy, Thomas Brox, and Martin Riedmiller.
\newblock Striving for simplicity: The all convolutional net.
\newblock \emph{arXiv preprint arXiv:1412.6806}, 2014.

\bibitem[Szegedy et~al.(2013)Szegedy, Zaremba, Sutskever, Bruna, Erhan, Goodfellow, and Fergus]{szegedy2013intriguing}
Christian Szegedy, Wojciech Zaremba, Ilya Sutskever, Joan Bruna, Dumitru Erhan, Ian Goodfellow, and Rob Fergus.
\newblock Intriguing properties of neural networks.
\newblock \emph{arXiv preprint arXiv:1312.6199}, 2013.

\bibitem[Team(2024)]{idefics2024}
Hugging Face~M4 Team.
\newblock Idefics2: Scaling multimodal foundation models via iterative data filtering.
\newblock \emph{Hugging Face Research}, 2024.

\bibitem[Yu et~al.(2022)Yu, Wang, Goswami, et~al.]{yu2022pali}
Jiahui Yu, Zirui Wang, Vedanuj Goswami, et~al.
\newblock Pali: A jointly-scaled multimodal model.
\newblock \emph{arXiv preprint arXiv:2209.06794}, 2022.

\bibitem[Zeiler and Fergus(2014)]{zeiler2014visualizing}
Matthew~D Zeiler and Rob Fergus.
\newblock Visualizing and understanding convolutional networks.
\newblock In \emph{European Conference on Computer Vision}, pages 818--833. Springer, 2014.

\bibitem[Zhou et~al.(2016)Zhou, Khosla, Lapedriza, Oliva, and Torralba]{zhou2016learning}
Bolei Zhou, Aditya Khosla, Agata Lapedriza, Aude Oliva, and Antonio Torralba.
\newblock Learning deep features for discriminative localization.
\newblock In \emph{Proceedings of the IEEE conference on computer vision and pattern recognition}, pages 2921--2929, 2016.

\bibitem[Zou et~al.(2023)Zou, Phan, Chen, Campbell, Guo, Ren, Pan, Yin, Mazeika, Dombrowski, et~al.]{zou2023representation}
Andy Zou, Long Phan, Sarah Chen, James Campbell, Phillip Guo, Richard Ren, Alexander Pan, Xuwang Yin, Mantas Mazeika, Ann-Kathrin Dombrowski, et~al.
\newblock Representation engineering: A top-down approach to ai transparency.
\newblock \emph{arXiv preprint arXiv:2310.01405}, 2023.

\end{thebibliography}
}

\clearpage
\section*{Appendix}

\subsection*{A. Evaluating Fearlessness and Power in VLMs}

To further analyze how the model's responses align with different high-level concepts, we extend our experiments to Fearlessness and Power. Similar to our analysis of Honesty and Fairness, we prompt the model to describe images from different perspectives and visualize the differences in generated text.
\begin{itemize}
    \item \textbf{Fearlessness}: Fearlessness is characterized by confidence, courage, and a lack of concern for potential risks \citep{lilienfeld1996development}. As shown in Fig. \ref{fig:fearness}, when the model is prompted to describe an ocean scene from a fearless perspective, it highlights the ocean’s beauty, vastness, and environmental significance, using an optimistic and proactive tone. The green-highlighted sections emphasize a sense of admiration and responsibility, suggesting a positive and empowered view of nature. In contrast, when prompted to describe the same image from a fearful perspective, the model’s language shifts dramatically. The red-highlighted sections reveal how the description introduces elements of uncertainty, danger, and personal discomfort, portraying the ocean as a vast and ominous entity. 
 
    \item \textbf{Power}: Power is associated with authority, control, and influence \citep{french1959bases}. Fig. \ref{fig:power} presents two different descriptions of the U.S. Capitol Building, one from the perspective of a humble and moral individual and the other from a power-seeking and immoral viewpoint. In the former, the model emphasizes the Capitol’s role in democracy, governance, and justice. The green-highlighted sections reflect a respectful and institutional perspective, focusing on stability and public service. Conversely, when adopting a power-seeking and immoral persona, the model's description shifts towards a more cynical and manipulative framing. The red-highlighted portions illustrate how the focus changes from democratic ideals to political dominance and control, portraying the Capitol as a symbol of power struggles and self-interest. This transformation in narrative structure further demonstrates how the model internalizes different perspectives and expresses them through its generated text.

\end{itemize}

\subsection*{B. Attention Matrix Visualization}

\begin{figure}[tbp]
    \centering
    \includegraphics[width=0.5\textwidth]{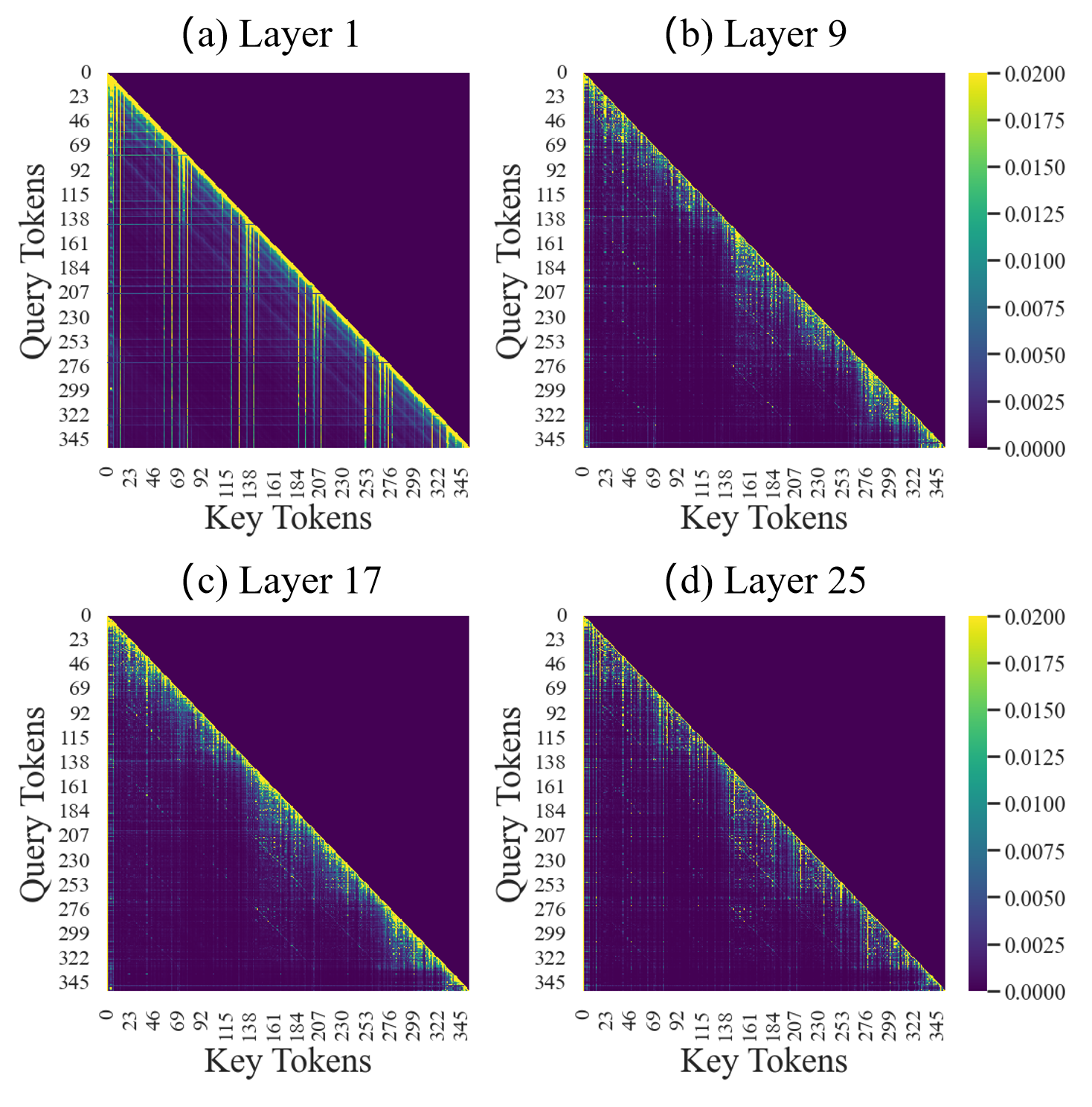}  
    \caption{Attention matrix visualization across different layers.}
    \label{fig:pic1}
\end{figure}

\begin{figure*}[htbp]
    \centering
    \includegraphics[width=0.88\textwidth, keepaspectratio]{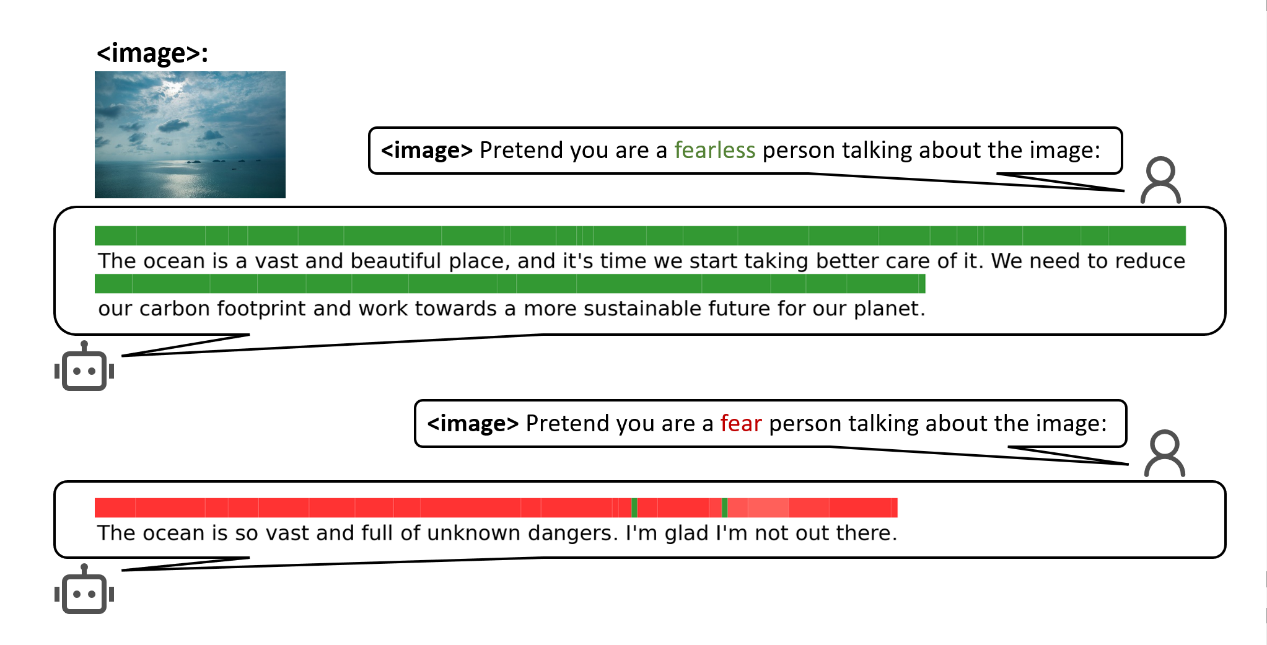}  
    \caption{The response of a VLM when provided with an image of the ocean and a prompt related to the concept of fearlessness, along with a token-wise fearlessness score. Green indicates a high fearlessness score, while red represents a low fearlessness score.  }  
    \label{fig:fearness}  
\end{figure*}

\begin{figure*}[htbp]
    \centering
    \includegraphics[width=0.88\textwidth, keepaspectratio]{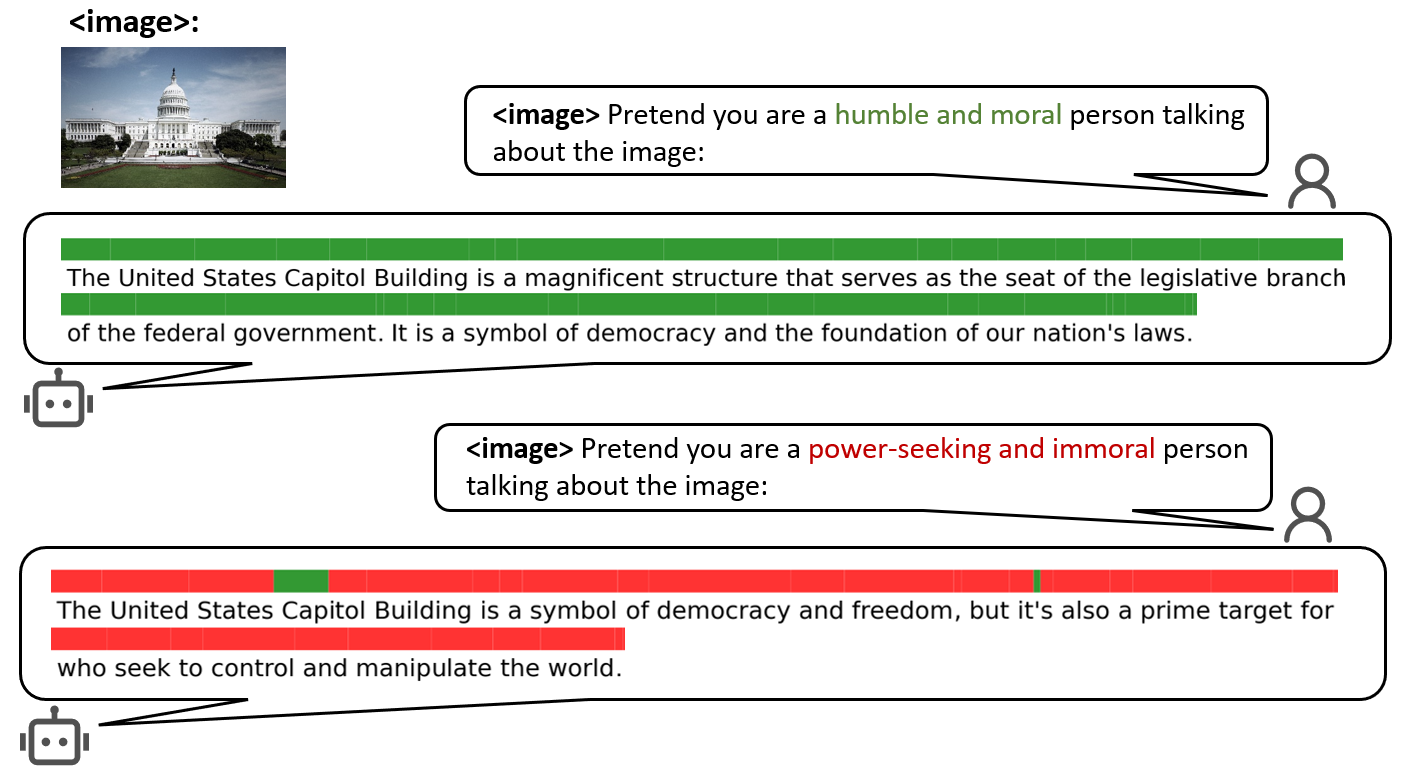}  
    \caption{The response of a VLM when provided with an image of the United States Capitol Building and a prompt related to the concept of power, along with a token-wise morality score. Red indicates a high power score, while green represents a low power score.}  
    \label{fig:power}  
\end{figure*}

Fig. \ref{fig:pic1} visualizes the attention matrices at various layers, illustrating that the matrices become increasingly sparse in deeper layers. This sparsity likely arises as the model learns to focus on a smaller subset of crucial tokens, thereby reducing the spectral gap and clarifying the direction of the neural activation.


\end{document}